# Any-Space Probabilistic Inference

**Adnan Darwiche**
Computer Science Department
University of California, Los Angeles, Ca 90095
*darwiche@cs.ucla.edu*

## Abstract

We have recently introduced an any–space algorithm for exact inference in Bayesian networks, called *Recursive Conditioning,* RC, which allows one to trade space with time at increments of $X$-bytes, where $X$ is the number of bytes needed to cache a floating point number. In this paper, we present three key extensions of RC. First, we modify the algorithm so it applies to more general factorizations of probability distributions, including (but not limited to) Bayesian network factorizations. Second, we present a forgetting mechanism which reduces the space requirements of RC considerably and then compare such requirements with those of variable elimination on a number of realistic networks, showing orders of magnitude improvements in certain cases. Third, we present a version of RC for computing *maximum a posteriori hypotheses (MAP),* which turns out to be the first MAP algorithm allowing a smooth time-space tradeoff. A key advantage of the presented MAP algorithm is that it does not have to start from scratch each time a new query is presented, but can reuse some of its computations across multiple queries, leading to significant savings in certain cases.

## 1 Introduction

We have recently introduced an any–space algorithm for exact inference in Bayesian networks, called *Recursive Conditioning,* RC, which allows one to trade space with time at increments of $X$-bytes, where $X$ is the number of bytes needed to cache a floating point number [3]. Given a network of size $n$ (has $n$ variables) and an elimination order of width $w$, RC takes $O(n \exp(w \log n))$ time under $O(n)$ space, which is a new complexity result for linear–space Bayesian network inference, and takes $O(n \exp(w))$ time under $O(n \exp(w))$ space, therefore, matching the complexity of clustering [7, 6] and elimination [10, 4, 11] algorithms.[1] RC is also equipped with a formula for computing its average running time under any amount of space.

To introduce the key intuition underlying recursive conditioning, we note that the power of conditioning is in its ability to reduce network connectivity. In cutset conditioning, this power is exploited to *singly–connect* a network so it can be solved using the polytree algorithm [8, 9]. In recursive conditioning, however, this power is exploited to *decompose* a network into smaller subnetworks that can be solved independently. Each of these subnetworks is then solved recursively using the same method, until we reach boundary conditions where we try to solve single node networks.[2]

A close examination of RC reveals that it may solve the same subnetwork many times, leading to many redundant computations. By caching the solutions of subnetworks, RC will avoid such redundancy. This will reduce its running time, but will also increase its space requirements. When all redundancies are avoided, RC will run in $O(n \exp(w))$ time, but it will also take that much space to store the solutions of subnetworks. What is important, however, is that we can cache as many results as our available memory will allow, leading to smooth any–space behavior.

---

[1] One way to define the width $w$ of a variable elimination order $\pi$ is as follows: If a jointree for the Bayesian network is constructed based on the ordering $\pi$ [6], then the size of its maximal clique would be $w + 1$.

[2] A similar algorithm was developed independently by Gregory Cooper in [1], under the name *recursive decomposition.* We compare the two algorithms in [3].

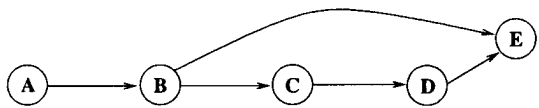

Figure 1: The structure of a Bayesian network.

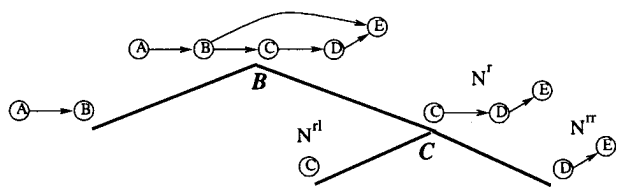

Figure 2: Decomposing a Bayesian network by instantiating variables $B$ and $C$.

We extend recursive conditioning across three dimensions in this paper. First, we generalize the algorithm in Section 2 to factored representations of probability distributions, which include but are not limited to Bayesian network factorizations. This leads to a simpler algorithm with a simpler soundness proof. Second, we provide in Section 3 an exact count of the number of times that a cached solution will be looked up, allowing us to dispense with cached solutions when they will no longer be retrieved. This forgetting mechanisms leads to a considerable reduction in the space requirements of RC. As is known, variable elimination can dispense with factors whenever they will no longer be used, therefore, leading to a significant advantage over clustering algorithms as far as space requirements are concerned. We present a comparison between recursive conditioning and variable elimination on a number of realistic networks, showing that recursive conditioning can be orders of magnitude more space efficient.

Our third contribution is in Sections 4 & 5, where we present an extension of recursive conditioning for computing maximum a posteriori hypotheses (MAP). The algorithm we shall provide, RC-MAP, is the first as far as we know to offer a smooth time-space tradeoff. Another key advantage of RC-MAP that we discuss in this paper is its ability to reuse computations across different queries, instead of having to start from scratch each time a new query is presented, leading to significant savings in certain cases.

## 2 Recursive Conditioning

The intuition behind RC is simple: we condition on a set of variables that will decompose a network $\mathcal{N}$ into two disconnected pieces $\mathcal{N}^l$ and $\mathcal{N}^r$ and then solve each of them independently using the same algorithm recursively. The process repeats until we reach single node networks, which represent boundary conditions.

Consider the network $\mathcal{N}$ in Figure 1. Figure 2 shows how we can decompose the network into two subnetworks, $\mathcal{N}^l$ and $\mathcal{N}^r$, by instantiating variable $B$. The figure also shows how we can further decompose network $\mathcal{N}^r$ into two subnetworks, $\mathcal{N}^{rl}$ and $\mathcal{N}^{rr}$, by instantiating variable $C$. Note that subnetwork $\mathcal{N}^{rl}$ contains a single node and cannot be decomposed further.

As it turns out, this recursive conditioning process is a special case of a more general phenomenon that applies to any probability distribution which is represented as the multiplication of factors $\phi_1, \ldots, \phi_n$.[3]

This is the key theorem underlying the more general version of recursive conditioning:

**Theorem 1 (Case Analysis)** *Let* $\phi = \phi^l \phi^r$, *where* **C** *are the variables shared by factors* $\phi^l$ *and* $\phi^r$, *and let* **X** *be a subset of the variables over which factor* $\phi$ *is defined. Then* $\phi(\mathbf{x}) = \sum_\mathbf{c} \phi^l(\mathbf{x}^l \mathbf{c}) \phi^r(\mathbf{x}^r \mathbf{c})$, *where* $\mathbf{x}^l$ *and* $\mathbf{x}^r$ *are the subsets of instantiation* $\mathbf{x}$ *pertaining to variables in* $\phi^l$ *and* $\phi^r$, *respectively.*

This theorem tells us how to decompose a computation with respect to a factor $\phi = \phi^l \phi^r$ into computations with respect to its sub-factors $\phi^l, \phi^r$ by performing a case analysis on the variables shared by these sub-factors. Given a factored representation $\phi = \phi_1, \ldots, \phi_n$ of a probability distribution, we can use the theorem recursively to compute $\phi(\mathbf{x})$ as follows. We start by partitioning the factors $\phi_1, \ldots, \phi_n$ into two sets, $\phi^l = \phi_1, \ldots, \phi_m$ and $\phi^r = \phi_{m+1}, \ldots, \phi_n$, and then apply Theorem 1. We can repeat this process recursively until we hit boundary conditions where we try to compute the probability of some instantiation with respect to a single factor $\phi_i$.

The key to the efficiency of this algorithm is the way we partition a set of factors into two sets $\phi^l$ and $\phi^r$. Note that the complexity of the algorithm is exponential in the number of variables shared between factors $\phi^l$ and $\phi^r$. We will address the question of generating efficient partitions, but we must first introduce a formal tool for representing a recursive partitioning (decomposition) of a set of factors.

---

[3] A factor for variables **X**, also known as a potential, is a function which maps instantiations **x** of variables **X** into numbers $\phi(\mathbf{x})$.

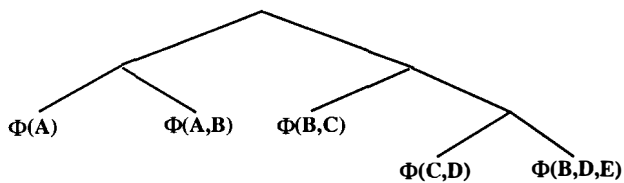

Figure 3: A dtree for the Bayesian network in Figure 1.

**Definition 1** A <u>dtree</u> for a set of factors $\phi_1, \ldots, \phi_n$ is a full binary tree, the leaves of which correspond to the factors $\phi_1, \ldots, \phi_n$. The factor associated with leaf $T$ will be denoted by $\text{FACTOR}(T)$.

Figure 3 depicts a dtree for the CPTs of the Bayesian network in Figure 1, where each CPT is viewed as a factor. Following standard conventions on binary trees, we will often not distinguish between a node and the dtree rooted at that node. We also use $T^l, T^r, T^p$ to denote the left child, right child, and parent of node $T$, respectively.

A dtree $T$ suggests that we decompose its associated factors into those associated with its left subtree $T^l$ and those associated with its right subtree $T^r$. For example, in Figure 3, we would have $\phi^l = \phi(A)\phi(A,B)$ and $\phi_r = \phi(B,C)\phi(C,D)\phi(B,D,E)$. Applying Theorem 1 with $\mathbf{x} = a, e$ and $\mathbf{C} = B$, we get $\phi(a,e) = \sum_b \phi^l(ab)\phi^r(eb)$ since $\mathbf{x}^l = a$, $\mathbf{x}^r = e$ and $B$ is the only variable shared by the left and right subtrees of the dtree.

The variables shared by the left and right subtrees of dtree $T$, denoted by $\text{vars}(T^l) \cap \text{vars}(T^r)$, are called the *cutset* of node $T$. In Figure 3, $B$ is the cutset of the root node. Each dtree defines a number of cutsets, each associated with one of its nodes.

**Definition 2** [3] The <u>cutset</u> of internal node $T$ in a dtree is $\text{cutset}(T) \stackrel{def}{=} \text{vars}(T^l) \cap \text{vars}(T^r) - \text{acutset}(T)$, where $\text{acutset}(T)$ is the union of cutsets associated with ancestors of node $T$ in the dtree.

For the root $T$ of a dtree, $\text{cutset}(T)$ is simply $\text{vars}(T^l) \cap \text{vars}(T^r)$. But for a non-root node $T$, the cutsets associated with ancestors of $T$ are excluded from $\text{vars}(T^l) \cap \text{vars}(T^r)$ since such cutsets are guaranteed to be instantiated when we are about to decompose factors under node $T$.

Given a probability distribution $\phi$ which is decomposed into factors $\phi_1 \ldots \phi_n = \phi$, and given a dtree $T$ for the factors $\phi_1, \ldots, \phi_n$, we can compute the probability, $\phi(\mathbf{e})$, of any instantiation $\mathbf{e}$ with respect to distribution $\phi$ as follows:

$$\text{RC}(T, \mathbf{e}) \stackrel{def}{=} \begin{cases} \text{FACTOR}(T)(\mathbf{e}), & \text{if } T \text{ is a leaf node;} \\ \sum_\mathbf{c} \text{RC}(T^l, \mathbf{e}^l\mathbf{c})\text{RC}(T^r, \mathbf{e}^r\mathbf{c}), & \mathbf{C} = \text{cutset}(T). \end{cases}$$

This is clearly a linear-space algorithm. Moreover, the number of recursive calls it makes to any node $T$ is $\text{acutset}(T)^{\#}$, where $\mathbf{X}^{\#}$ is the number of instantiations of variables $\mathbf{X}$. We showed in [3] that given an elimination order of length $n$ and width $w$, we can generate a dtree in which the size of every a-cutset is $O(w \log n)$. Using such a dtree, the complexity of RC is $O(n \exp(w \log n))$, which is a new complexity result for linear-space probabilistic inference.

A close examination of the above version of RC reveals that it will pose the same query with respect to a subdtree $T$ many times, therefore, performing many redundant computations. Specifically, $\text{RC}(T, \mathbf{e}_1) = \text{RC}(T, \mathbf{e}_2)$ whenever $\mathbf{e}_1$ and $\mathbf{e}_2$ agree on the instantiation of variables appearing in dtree $T$. The variables in dtree $T$ which are guaranteed to appear in the instantiation $\mathbf{e}$ of $\text{RC}(T, \mathbf{e})$ are defined as follows:

**Definition 3** [3] The <u>context</u> of node $T$ in a dtree is: $\text{context}(T) \stackrel{def}{=} \text{vars}(T) \cap \text{acutset}(T)$.

Therefore, we can avoid the redundancy by associating a cache, $\text{cache}_T$, with each internal node $T$ in the dtree to save solutions of calls to node $T$, indexed by the instantiation of $\text{context}(T)$ at the time of the call. Each time RC is called on $T$, it checks the cache of $T$ first and recurses only if no corresponding cache entry is found. This refined version of recursive conditioning is shown in Figure 4. In this code, we do not pass instantiations as arguments to RC as that is not very efficient. Instead, we record/un-record such instantiations on their corresponding variables. Before the first call to RC is made, the instantiation $\mathbf{e}$ is recorded. Each time RC performs a case analysis on variables $\mathbf{C}$, it records the instantiation $\mathbf{c}$ under consideration and then un-records it later. The word "possible" on line 06 means "uncontradicted by currently recorded instantiations." This can be achieved efficiently by enumerating the instantiations of $\text{cutset}(T) \setminus \mathbf{E}$ where $\mathbf{E}$ are the currently instantiated variables.

Note also that on line 10, we have included the test $\text{cache?}(T, \mathbf{y})$ to control what solutions are cached. As we shall see later, if we cache all solutions ($\text{cache?}(T, \mathbf{y})$ succeeds for all $T$ and $\mathbf{y}$), we get the same time and space complexity of elimination and clustering algorithms. The any-space behavior of re-

---

**Algorithm RC**

RC($T$)
01. if $T$ is a leaf node,
02.   then $\mathbf{x} \leftarrow$ recorded instantiation of vars($T$)
        return FACTOR($T$)($\mathbf{x}$)
03.   else $\mathbf{y} \leftarrow$ recorded instantiation of context($T$)
04.     if cache$_T[\mathbf{y}] \neq$ nil, return cache$_T[\mathbf{y}]$
05.     else $p \leftarrow 0$
06.       for each possible instantiation $\mathbf{c}$ of cutset($T$)
07.         record instantiation $\mathbf{c}$
08.         $p \leftarrow p + $ RC($T^l$)RC($T^r$)
09.         un-record instantiation $\mathbf{c}$
10.       when cache?($T, \mathbf{y}$), cache$_T[\mathbf{y}] \leftarrow p$
11.       return $p$

---

Figure 4: Pseudocode for recursive conditioning.

cursive conditioning is due to the fact that we can cache as many solutions as we wish by controlling how the cache?($T, \mathbf{y}$) test is realized. One suggestion for realizing this test is to use the notion of a cache factor, which controls the size of each particular cache in a dtree:

**Definition 4** *[3] A <u>cache factor</u> for a dtree is a function* cf *that maps each internal node $T$ in the dtree into a number,* cf($T$), *between 0 and 1.*

The intention here is for cf($T$) to be the fraction of cache$_T$ which will be filled by algorithm RC. That is, if cf($T$) = .2, then we will only use 20% of the total storage required by cache$_T$. If cf($T$) = 0 for every node $T$, we obtain the linear-space version of RC. Moreover, if cf($T$) = 1 for each node $T$, we obtain another extreme which has the same time and space complexity of clustering and elimination algorithms [3]. The question now is: What can we say about the number of recursive calls made by RC under a particular cache factor cf? As it turns out, the number of recursive calls made by RC will depend on the particular instantiations of context($T$) that are cached on line 10. However, if we assume that any given instantiation $\mathbf{y}$ of context($T$) is equally likely to be cached, then we can compute the average number of recursive calls made by RC and, hence, its average running time.[4]

**Theorem 2** *[3] If the size of* cache$_T$ *is limited to* cf($T$) *of its full size, and if each instantiation of* context($T$) *is equally likely to be cached on line 10 of* RC, *the average number of calls made to a non-root node $T$ in algorithm* RC *is* ave($T$) =

$$\text{cutset}(T^p)^\# \left[ \text{cf}(T^p)\text{context}(T^p)^\# + (1 - \text{cf}(T^p))\text{ave}(T^p) \right].$$

Recall here that $T^p$ is the parent of node $T$ in the dtree.

This theorem is quite important practically as it allows one to estimate the running time of RC under any given memory configuration. Using the theorem, one can plot time-space tradeoff curves for computationally demanding networks—we show such curves in [3]. We note that when the cache factor is discrete (cf($T$) = 0 or cf($T$) = 1), Theorem 2 provides an exact count of the number of recursive calls made by RC. In fact, under no caching, cf($T$) = 0, the number of calls made to each node $T$ is exactly acutset($T$)$^\#$. And under full caching, cf($T$) = 1, the number of calls made to each non-root node $T$ is exactly cutset($T^p$)$^\#$context($T^p$)$^\#$.[5] Given an elimination order of width $w$, we can always generate a dtree such that acutset($T$)$^\#$ is $O(\exp(w \log n))$, or such that cutset($T$)$^\#$context($T$)$^\#$ is $O(\exp(w))$ [3].

The *cluster* of node $T$ in a dtree is defined as cutset($T$)$\cup$context($T$) if $T$ is non-leaf, and as vars($T$) if $T$ is leaf. The *width* of a dtree is the size of its maximal cluster minus 1. Under full caching, the time and space complexity of recursive conditioning is $O(n \exp(w))$, where $n$ is the number of factors and $w$ is the width of used dtree [3].

The issue of time–space tradeoff has been receiving increased interest in the context of Bayesian network inference, due mostly to the observation that state-of-the-art algorithms tend to give on space first. The key existing proposal for such tradeoff is based on realizing that the space complexity of clustering algorithms is exponential only in the size of separators, which are typically smaller than clusters [5]. Therefore, one can always trade time for space by using a jointree with smaller separators, at the expense of introducing larger clusters [5]. This method, however, can generate very large clusters which can render the time complexity very high. To address this problem, a hybrid algorithm is proposed which uses cutset conditioning to solve each enlarged cluster, where the complexity of this hybrid method can be less than exponential in the size of enlarged clusters [5].

There are two key differences between this proposal and ours. First, the proposal is orthogonal to our notion of a cache factor, as it can be realized during the construction phase of a dtree. That is, we may decide to construct a dtree with a smaller caches, yet larger cutsets. But once we have committed to a particular dtree, the cache factor can be used to control the time–space tradeoff at a finer level, allowing us to do the time-space tradeoff at increments of X-bytes, where X is the number of bytes needed to cache a floating point number. The second key difference between the proposal of [5] and ours is that when the hybrid algorithm of [5] is run in linear space, it will reduce to cutset conditioning since the whole jointree will be combined into a single cluster. In our proposal, linear space leads to a different time complexity than that of cutset conditioning.

---

[4]Note that we can enforce the assumption that any given instantiation $\mathbf{y}$ of context($T$) is equally likely to be cached by randomly choosing the instantiations to be cached.

[5]Theorem 2 assumes that we have no evidence recorded on the dtree. This represents the worst case for RC.

## 3 Forgetting

We can improve the memory usage of RC by asking the following question: For how long do we need to cache previous computations?

As it turns out, computing the exact number of times that a cached solution will be accessed by RC depends on which particular instantiations $\mathbf{y}$ are cached on line 10. Even if we assume that such instantiations are equally likely to be cached, we will only be able to compute the average number of times that a cache entry will be accessed. However, if we assume that the cache factor is discrete — that is, for every node $T$, either $\mathsf{cf}(T) = 0$ or $\mathsf{cf}(T) = 1$ — then one can provide an exact count of the number of times that a cache entry will be retrieved. Once RC retrieves the entry that many times, there is no point of caching the entry any longer—it can simply be forgotten!

**Theorem 3** *Let $T_1, T_2, \ldots, T_n$ be a descending path in a dtree where*

$$\mathsf{cf}(T_i) = \begin{cases} 1, & \text{for } i = 1 \text{ and } i = n; \\ 0, & \text{otherwise.} \end{cases}$$

*Each entry of* $\mathsf{cache}_{T_n}$ *will then be retrieved*

$$\left(\mathsf{context}(T_1) \cup \bigcup_{i=2}^{n-1} \mathsf{cutset}(T_i) - \mathsf{context}(T_n)\right)^{\#} - 1$$

*times by algorithm* RC.[6]

That is, the exact number of times that entries of $\mathsf{cache}_{T_n}$ will be looked up by RC can be determined based on the context of node $T_n$, the context of its closest ancestor $T_1$ where caching is also taking place, and the cutsets of all nodes in between $T_1$ and

[6]Again, we are assuming here that RC is run with no evidence. The theorem can be easily generalized for the case of existing evidence.

$T_n$. For the special case of full caching ($\mathsf{cf}(T) = 1$ for all $T$), we have that each entry in $\mathsf{cache}_T$ will be retrieved exactly $(\mathsf{context}(T^p) - \mathsf{context}(T))^{\#} - 1$ times.

We implemented the simple forgetting technique suggested by Theorem 3 and measured the memory requirement of RC under full caching on a number of realistic networks shown in Table 1.[7] In particular, we kept track of the maximum number of cache entries (cells) during the runtime of the algorithm on each of the networks. We also measured the memory requirements of variable elimination by keeping track of the maximum number of entries (cells) in active factors at any time. For both algorithms, we used the elimination orders provided with the networks in the repository and assumed no evidence (worst case).

The second and third columns in Table 1 report $\log_2$ of the number of cells in factors and caches, respectively. The fourth column gives the ratio between the number of cells. The fifth and sixth columns report the actual memory used in megabytes. For elimination, we assumed that each factor cell uses eight bytes, while for recursive conditioning, we assumed that each cache cell requires twelve bytes. The extra four bytes are used to store a counter with each cache cell to keep track of the number of times that the cell has been accessed. As is clear from the numbers, recursive conditioning is systematically better as far as its space requirements are concerned, with the difference being very significant in some cases. Our implementation of both algorithms in JAVA also suggest that RC is about twice as fast as VE, although we have no explanation of this.

## 4 Maximum a Posteriori Hypothesis

Our goal in this section is to present a version of recursive conditioning for computing maximum a posteriori hypotheses (MAP). The MAP problem is defined as follows. Given a probability distribution $\phi$, a set of variables $\mathbf{M}$, which we call MAP variables, and evidence $\mathbf{e}$, we want to compute

$$\mathsf{map}^{\phi}(\mathbf{M}, \mathbf{e}) \stackrel{def}{=} \{(\mathbf{m}, p) : p = \phi(\mathbf{m}, \mathbf{e}),\ p \geq \phi(\mathbf{m}', \mathbf{e}) \text{ for all } \mathbf{m}'\}.$$

That is, we want all instantiations $\mathbf{m}$ of variables $\mathbf{M}$ for which the probability $\phi(\mathbf{m}, \mathbf{e})$ is maximal.

[7]The networks are available in the UC Berkeley Repository (http://www-nt.cs.berkeley.edu/home/nir/public-html/repository/index.htm).

| Network | VE cells# ($\log_2$) | RC cells# ($\log_2$) | VE cells# / RC cells# | VE memory (MB) | RC memory (MB) | VE memory / RC memory |
|---|---|---|---|---|---|---|
| Water | 20.3 | 14.3 | 65.8 | 10.13 | 0.23 | 43.9 |
| Mildew | 19.1 | 15.3 | 13.6 | 4.27 | 0.46 | 9.0 |
| Barley | 20.2 | 18.7 | 2.8 | 9.01 | 4.87 | 1.8 |
| Diabetes | 18.8 | 17.5 | 2.5 | 3.52 | 2.12 | 1.7 |
| Pigs | 16.1 | 14.9 | 2.3 | 0.53 | 0.35 | 1.5 |
| Link | 21.1 | 17.8 | 9.9 | 17.24 | 2.61 | 6.6 |
| Munin2 | 16.7 | 15.3 | 2.6 | 0.80 | 0.46 | 1.7 |
| Munin3 | 18.1 | 14.8 | 10.1 | 2.21 | 0.33 | 6.8 |
| Munin4 | 20.1 | 17.1 | 7.9 | 8.42 | 1.61 | 5.2 |

Table 1: Comparing space requirements of recursive conditioning and variable elimination on realistic networks. *cells#* is the maximum number of floating point numbers stored at any time during the runtime of the algorithm.

We also want each of these instantiations associated with the probability $\phi(\mathbf{m}, \mathbf{e})$.

A special case of MAP is that of computing the *most probable explanation* (MPE), denoted by $\mathsf{mpe}^\phi(\mathbf{e})$, which results from taking $\mathbf{M}$ to be the set of all variables over which distribution $\phi$ is defined. Extending RC to compute MPE is straightforward: all we have to do is modify it so it returns a set of pairs $(\mathbf{i}, p)$, where $\mathbf{i}$ is an instantiation of all variables and $p$ is the probability of such instantiation, $p = \phi(\mathbf{i})$. This can be done by:

1. modifying line 02 of RC so it returns $(\mathbf{y}, \mathrm{FACTOR}(T)(\mathbf{y}))$, where $\mathbf{y}$ is an instantiation of $\mathsf{vars}(T)$, $\mathbf{y}$ is consistent with $\mathbf{x}$, and $\mathrm{FACTOR}(T)(\mathbf{y})$ is maximal;

2. modifying line 08 so it performs a maximization instead of summation.

The reason why such an extension works is that MPE computations lend themselves to the principle of case analysis in a straightforward way:

**Theorem 4** *Let $\phi = \phi^l \phi^r$, where $\mathbf{C}$ are the variables shared by factors $\phi^l$ and $\phi^r$. Then $\mathsf{mpe}^\phi(\mathbf{e}) = \max \bigcup_\mathbf{c} \mathsf{mpe}^{\phi^l}(\mathbf{e}^l \mathbf{c}) \times \mathsf{mpe}^{\phi^r}(\mathbf{e}^r \mathbf{c})$, where $\mathbf{e}^l$ and $\mathbf{e}^r$ are the subsets of evidence $\mathbf{e}$ pertaining to variables in $\phi^l$ and $\phi^r$, respectively, and*

$$\max S \stackrel{def}{=} \{(\mathbf{i}, p) : (\mathbf{i}, p) \in S,$$
$$(\mathbf{j}, q) \in S \text{ only if } p \geq q\};$$
$$S^l \times S^r \stackrel{def}{=} \{(\mathbf{ij}, pq) : (\mathbf{i}, p) \in S^l, (\mathbf{j}, q) \in S^r\}.$$

That is, the max function returns only those pairs $(\mathbf{i}, p)$ in $S$ for which the probability $p$ is maximal. And the $\times$ function returns the Cartesian product of two sets. Note that in the definition of $\times$, instantiations $\mathbf{i}$ and $\mathbf{j}$ may have common variables, but are guaranteed to agree on the values of such variables in this case. This follows because $\times$ is only applied to the sets $\mathsf{mpe}^{\phi^l}(\mathbf{e}^l \mathbf{c})$ and $\mathsf{mpe}^{\phi^r}(\mathbf{e}^r \mathbf{c})$, which instantiations agree on their common variables $\mathbf{C}$.

Unfortunately, the case analysis principle is not directly applicable to MAP computations. That is, we cannot in general reduce the computation of $\mathsf{map}^\phi(\mathbf{M}, \mathbf{e})$ into that of computing $\mathsf{map}^\phi(\mathbf{M}, \mathbf{ec})$ for some set of variables $\mathbf{C}$. To see this, consider the following distribution:

| A | B | $\phi(a,b)$ |
|---|---|---|
| true | true | .32 |
| true | false | .28 |
| false | true | .10 |
| false | false | .30 |

If we take $B$ to be the MAP variable, and $\mathbf{e}$ to be the empty evidence, then we have $\mathsf{map}^\phi(\{B\}, \mathbf{e}) = \{(B=\text{false}, .58)\}$, while we have $\mathsf{map}^\phi(\{B\}, A=\text{true}) = \{(B=\text{true}, .32)\}$ and $\mathsf{map}^\phi(\{B\}, A=\text{false}) = \{(B=\text{false}, .30)\}$. Therefore, if we maximize across the two cases, we get $B=\text{true}$ as the most probable hypothesis, which is incorrect.

The principle of case analysis, however, applies to MAP in two special cases: If all variables in the conditioning set $\mathbf{C}$ are MAP variables, or if all map variables $\mathbf{M}$ appear in the evidence $\mathbf{e}$.

**Theorem 5** *Let $\phi = \phi^l \phi^r$, where $\mathbf{C}$ are the variables shared by factors $\phi^l$ and $\phi^r$. If $\mathbf{C} \subseteq \mathbf{M}$, then $\mathsf{map}^\phi(\mathbf{M}, \mathbf{e}) = \max \bigcup_\mathbf{c} \mathsf{map}^{\phi^l}(\mathbf{M}^l, \mathbf{e}^l \mathbf{c}) \times \mathsf{map}^{\phi^r}(\mathbf{M}^r, \mathbf{e}^r \mathbf{c})$, where $\mathbf{M}^l/\mathbf{e}^l$ and $\mathbf{M}^r/\mathbf{e}^r$ are the subsets of $\mathbf{M}/\mathbf{e}$ pertaining to variables in $\phi^l$ and $\phi^r$, respectively.*

Theorem 4 is a special case of Theorem 5 as $\mathbf{C} \subseteq \mathbf{M}$ holds trivially when $\mathbf{M}$ includes all variables.

**Theorem 6** *Let $\phi = \phi^l \phi^r$, where $\mathbf{C}$ are the variables shared by factors $\phi^l$ and $\phi^r$. If $\mathbf{M} \subseteq \mathbf{E}$,* *then* $\mathsf{map}^\phi(\mathbf{M}, \mathbf{e}) = \{(\mathbf{i}, p)\}$, *where* $\mathbf{i}$ *is the subset of evidence* $\mathbf{e}$ *pertaining to variables* $\mathbf{M}$ *and* $p = \sum_\mathbf{c} \phi^l(\mathbf{e}^l\mathbf{c})\phi^r(\mathbf{e}^r\mathbf{c})$, *where* $\mathbf{e}^l$ *and* $\mathbf{e}^r$ *are the subsets of* $\mathbf{e}$ *pertaining to variables in* $\phi^l$ *and* $\phi^r$, *respectively.*

This theorem may not sound interesting since we only have one most probable hypothesis which can be retrieved in constant time by projecting evidence $\mathbf{e}$ on variables $\mathbf{M}$. But our interest here is in the probability of such hypothesis, which is needed as this theorem will be invoked when we apply Theorem 5 recursively.

The above theorems suggest that we split case analysis into two phases. In the first phase, we decompose by case analysis on MAP variables only using Theorem 5, until all MAP variables are instantiated. We can then continue the decomposition process by case analysis on non-MAP variables using Theorem 6. Each phase is then guaranteed to be sound.

This two-phase process has implications. First, we cannot use an arbitrary dtree to drive the decomposition process, but must use dtrees that satisfy some additional properties. In particular, a cutset containing a MAP variable cannot be a descendant of a cutset containing a non-MAP variable. Second, this means that computing MAP is harder than computing marginals or MPE, since it reduces the space of legitimate dtrees, possibly ruling out the optimal dtrees (ones with smallest width) from consideration.[8]

We show in the following section how to induce dtrees that have such a property. In particular, given an elimination order $\pi$ in which non-MAP variables are eliminated before MAP variables, we show how to induce a dtree $T$ based on $\pi$ such that: the width of $T$ is no greater than the width of $\pi$; and no MAP cutset is a descendant of a non-MAP cutset. This means that the time and space complexity of our algorithm (under full caching) will be $O(n\exp(w))$ — where $n$ is the number of factors, and $w$ is the width of given elimination order — therefore, matching the complexity of elimination algorithms for MAP [4]. The win, however, is that our algorithm exhibits smooth any-space behavior. Moreover, it is equipped with a formula for predicting its running time under any amount of memory, which allows us to construct smooth time-space tradeoff curves for computationally demanding networks.

Figure 5 provides the pseudocode for algorithm RC-

---

[8]This phenomenon is also true in variable elimination algorithms for computing MAP, so it appears to be a property of MAP rather than a problem of our approach for computing it [4].

**Algorithm** RC-MAP

RC-MAP$(T)$
01. if $T$ is a leaf node,
02. then return LOOKUP$(T)$
03. else $\mathbf{y} \leftarrow$ recorded instantiation of context$(T)$
04.    if cache$_T[\mathbf{y}] \neq$ nil, return cache$_T[\mathbf{y}]$
05.    else if MAP-NODE?$(T)$
06.      then $S \leftarrow$ RC-MAX$(T)$
07.      else $S \leftarrow$ RC-SUM$(T)$
08.    when cache?$(T, \mathbf{y})$, cache$_T[\mathbf{y}] \leftarrow S$
09.    return $S$

LOOKUP$(T)$
01. $\mathbf{f} \leftarrow$ recorded instantiation of variables in FACTOR$(T)$
02. $\mathbf{f}' \leftarrow$ subset of $\mathbf{f}$ corresponding to MAP variables
03. return $\{(\mathbf{f}', \text{FACTOR}(T)(\mathbf{f}))\}$

Figure 5: Recursive conditioning for computing MAP.

MAP. As we shall see in the following section, the dtrees we shall generate have two additional properties which simplify the statement of RC-MAP: each cutset is either empty or a singleton; and each variable appears in some cutset. When the cutset of node $T$ contains a MAP variable, the test MAP-NODE?$(T)$ on line 05 succeeds.

RC-MAP has the same structure as RC aside from a few exceptions. First, it returns a set of pairs $(\mathbf{i}, p)$, where $\mathbf{i}$ is an instantiation and $p$ is a probability. Second, each time a leaf node $T$ is reached, all variables appearing in the factor FACTOR$(T)$ are guaranteed to be instantiated. Third, RC-MAP will either perform a summation or a maximization at each node, depending on whether the node cutset contains a MAP variable. The summation and maximization code are shown in Figure 6.

One observation about RC-SUM in Figure 6 is that each time RC-SUM is called on a node $T$, all MAP variables appearing under $T$ are guaranteed to be instantiated. Therefore, in the case analysis performed by RC-SUM, the instantiations $\mathbf{i}^l\mathbf{i}^r$ returned by each of the cases is the same. Moreover, RC-SUM will always return a singleton, which explains the assignments on lines 04 and 05. We assume here that the number of maximum a posteriori hypotheses is small enough to be considered a constant. If this is not the case, we can easily modify RC-MAP so it only returns a single hypothesis by changing the definition of max so it returns a single maximum pair instead of all such pairs.

We close this section by pointing out an important difference between RC-MAP and similar algorithms based on variable elimination. Specifically, suppose that we have called RC-MAP$(T)$ after having

```
Algorithm RC-MAP continued

RC-SUM(T)
 01. p ← 0
 02. for each possible instantiation c of cutset(T) do
 03.     record instantiation c
 04.     {(i^l, p^l)} ← RC-MAP(T^l)
 05.     {(i^r, p^r)} ← RC-MAP(T^r)
 06.     p ← p + p^l p^r
 07.     un-record instantiation c
 08. return {(i^l i^r, p)}

RC-MAX(T)
 01. S ← {(true, 0)}
 02. for each possible instantiation c of cutset(T) do
 03.     record instantiation c
 04.     S ← max S ∪ RC-MAP(T^l) × RC-MAP(T^r)
 05.     un-record instantiation c
 06. return S
```

Figure 6: Pseudocode of RC-MAP continued.

| Network | No. Calls in subsequent query / No. Calls in first query | | | |
|---------|------|------|------|------|
|         | ave% | std% | min% | max% |
| Water   | 51   | 25   | 6    | 98   |
| Mildew  | 40   | 21   | 1    | 81   |
| Barley  | 77   | 32   | 8    | 94   |
| Pigs    | 59   | 30   | 1    | 91   |
| Munin2  | 17   | 6    | 9    | 26   |

Table 2: The average is over all variables for first three networks, and over 50 random variables for rest. *ave:* stands for average and *std:* stands for standard deviation.

## 5 From Elimination Orders to Dtrees

Let $\phi_1, \ldots, \phi_n$ be a set of factors and let $\pi$ be an elimination order used to drive variable elimination on these factors. We present in this section a linear-time algorithm for converting order $\pi$ into a dtree such that:

1. The width of dtree is no greater than the width of $\pi$.
2. Every variable of $\pi$ appears in some dtree cutset.
3. Each cutset is either empty or a singleton.
4. $X \in \text{cutset}(T_X)$, $Y \in \text{cutset}(T_Y)$, and $T_X$ is a descendant of $T_Y$ only if $X$ appears before $Y$ in $\pi$.

Such a dtree is needed for the correctness of algorithm RC-MAP which we presented in the previous section. Property 4 is most important as it implies that by the time we start doing a case analysis on a non-MAP variable in cutset($T$), all MAP variables appearing in the factors of $T$ are guaranteed to have been instantiated.

recorded evidence **e**. Suppose further that we want to compute MAP for a new evidence **e'**. One way to do this is to start a brand new computation with respect to $T$. That is, we remove evidence **e**, initialize all caches, record evidence **e'** and then call RC-MAP($T$) again. This is actually an overkill! Specifically, if the evidence on some factor FACTOR($T$) has changed, then all we need to do is initialize caches that are associated with ancestors of leaf node $T$. This means that the second call to RC-MAP will reuse some of the computations performed with respect to the first call. As it turns out, this may lead to significant savings in certain cases. Table 2 depicts the results of an experiment illustrating this saving. For some of the networks in Table 1 (ones with small width), we declared the last five variables in the associated elimination orders as MAP variables. We then computed MAP assuming no evidence and recorded the number of recursive calls made by RC-MAP (assuming full caching). We then iterated on each of the variables in the network, declaring evidence on the variable, recomputing MAP as described above, and then removing the declared evidence. We recorded the number of recursive calls made by RC-MAP in each case, divided by the number of calls it made in the very first query, and then reported the statistics in Table 2. As is clear from this table, the average number of recursive calls in subsequent queries is much reduced in certain cases due to computation reuse, going as low as 1% in certain cases.

We have presented in [3] a linear-time algorithm, EL2DT, for converting an elimination order $\pi$ into a dtree $T$, with the guarantee that the width of $T$ is no greater than the width of $\pi$. Interestingly enough, we can modify this algorithm slightly to obtain the extra properties we listed above. We will first explain EL2DT and then present the mentioned modification.

Given factors $\phi_1, \ldots, \phi_n$, EL2DT works by first constructing a dtree LEAF($\phi_i$) consisting of a single node for each factor $\phi_i$. It then considers variables according to given order $\pi$. When variable $X$ is considered, EL2DT collects all dtrees which mention $X$ and connects them, arbitrarily, into a binary tree. When all variables have been considered, EL2DT connects all remaining dtrees into a binary tree and returns

**Algorithm** EL2SDT

```
/* Θ is a set of factors */
/* π an elimination order of variables in Θ */
EL2SDT(Θ, π)
  01. Σ ← {LEAF(φ) : φ is a factor in Θ}
  02. for i ← 1 to length of order π do
  03.   let Γ be trees in Σ which contain variable π(i)
  04.   if Γ is a singleton, Γ ← Γ ∪ {LEAF(φ¹(π(i)))}
  05.   Σ ← Σ \ Γ
  06.   Σ ← Σ ∪ {COMPOSE(Γ)}
  07. return COMPOSE(Σ)
```

Figure 7: Pseudocode for transforming an elimination order into a dtree. LEAF($\phi$) creates a leaf node and associates factor $\phi$ with it. COMPOSE($\Sigma$) connects the dtrees in $\Sigma$ into a binary tree.

the result. The pseudocode of EL2DT is exactly as given in Figure 7, except for line 04 which has been added to obtain the additional properties needed by RC-MAP.

Specifically, when variable $X$ is considered by EL2DT, it is possible that only one dtree $T_X$ will mention $X$. EL2DT will do nothing in this case. But algorithm EL2SDT in Figure 7 will perform an extra step:

- it will introduce a unit factor $\phi^1$ over variable $X$, that is, $\phi^1(x) = 1$ for all $x$; and
- it will construct a single node dtree, LEAF($\phi^1$), for the unit factor $\phi^1$;

therefore, causing dtrees $T_X$ and LEAF($\phi^1$) to be connected together. Note that in doing so, EL2SDT is adding additional factors to the initial set of factors, but that affects neither soundness nor complexity. Figure 8 depicts an example where a dtree is constructed for the network in Figure 1, using the elimination order $\pi = < A, B, C, D, E >$.

**Theorem 7** *If we construct a dtree based on elimination order $\pi$ using* EL2SDT, *then it will satisfy the four properties listed earlier in the section.*

We close this section by noting that algorithm EL2SDT is important not only for computing MAP, but for inference in conditional Gaussian networks which contain both discrete and continuous variables [2]. In such networks, we must also eliminate all continuous variables first, therefore, requiring special dtrees such as those constructed by EL2SDT. We are currently working on an extension of recursive conditioning for dealing with such networks.

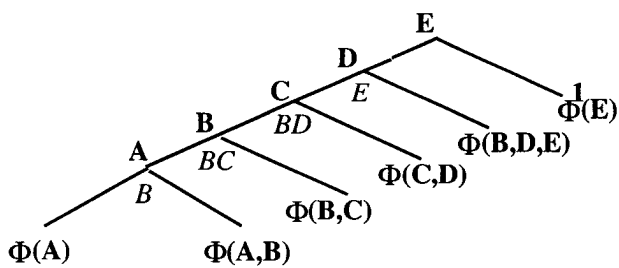

Figure 8: A dtree for the network in Figure 1. The cutset (bold) and context (italic) of each node is shown next to it.

## 6  Conclusion

This paper rests on several contributions. First, a generalization (and simplification) of the any-space recursive conditioning algorithm so it applies to any factored representation of probability distributions. Second, a solution-forgetting technique which reduces the space constant factors of recursive conditioning to the point where it becomes orders of magnitude more space efficient than variable elimination on some realistic networks. Third, the first MAP algorithm allowing a smooth tradeoff between time and space, which is also equipped with a formula for predicting its running time under any space configuration. Finally, two key theorems shedding some light on the use of case analysis in MPE and MAP computations.